%% file: paper.tex
\pdfoutput=1
\documentclass[10pt,twocolumn,letterpaper]{article}

\usepackage{iccv}
\makeatletter
\@namedef{ver@everyshi.sty}{}
\makeatother
\usepackage{tikz}

\usepackage{times}
\usepackage{epsfig}
\usepackage{graphicx}
\usepackage{amsmath}
\usepackage{amssymb}
\usepackage{amsthm}
\usepackage{subcaption}

\usepackage[accsupp]{axessibility}

\DeclareMathOperator{\Tr}{tr}
\DeclareMathOperator*{\argmax}{arg\,max}
\DeclareMathOperator*{\argmin}{arg\,min}

\theoremstyle{plain}
\newtheorem{theorem}{Theorem}
\newtheorem{lemma}[theorem]{Lemma}

\usepackage{parskip}

\usepackage[ruled,vlined]{algorithm2e}

\usepackage[pagebackref=true,breaklinks=true,letterpaper=true,colorlinks,bookmarks=false]{hyperref}

\iccvfinalcopy 

\def\iccvPaperID{9108} 

\ificcvfinal\fi

\begin{document}

\title{Rotation Averaging in a Split Second: \\ A Primal-Dual Method and a Closed-Form for Cycle Graphs}

\author{Gabriel Moreira \qquad
Manuel Marques \qquad
João Paulo Costeira \\
Institute for Systems and Robotics, Instituto Superior T\'ecnico\\
Av. Rovisco Pais, Lisboa, Portugal\\
{\tt\small \{gmoreira, manuel, jpc\}@isr.tecnico.ulisboa.pt} 
}

\maketitle
\ificcvfinal\thispagestyle{empty}\fi

\begin{abstract}
\vspace{-0.4cm}
A cornerstone of geometric reconstruction, rotation averaging seeks the set of absolute rotations that optimally explains a set of measured relative orientations between them. In spite of being an integral part of bundle adjustment and structure-from-motion, averaging rotations is both a non-convex and high-dimensional optimization problem. In this paper, we address it from a maximum likelihood estimation standpoint and make a twofold contribution. Firstly, we set forth a novel initialization-free primal-dual method which we show empirically to converge to the global optimum. Further, we derive what is to our knowledge, the first optimal closed-form solution for rotation averaging in cycle graphs and contextualize this result within spectral graph theory. Our proposed methods achieve a significant gain both in precision and performance.
\end{abstract}
\vspace{-0.2cm}

\section{Introduction}
Rotation averaging, also known as group synchronization, is an estimation problem wherein we want to find a set of rotations $\{R_1,\dots,R_n\} \in \mathrm{SO}(p)^n$, where
\begin{equation}
    \mathrm{SO}(p) = \{R \in \mathbb{R}^{p\times p} : RR^\top = I, \mathrm{det}(R)=1\},
\end{equation}
that optimally explains a set of $m$ noisy pairwise measurements $\{\widetilde{R}_{ij}\}_{i\sim j} \in \text{SO}(p)^m$ of the relative orientations $R_i R_j^\top$. The notation $i\sim j$ refers to the existence of a measurement between rotations $i$ and $j$. As a sub-problem of several 3D reconstruction tasks, namely bundle adjustment \cite{agarwal2010,zhang2020}, structure-from-motion \cite{Schonberger2016,Kasten2019} and camera network calibration \cite{Tron2009}, rotation averaging is of particular interest in computer vision. Nevertheless, the high-dimensionality of the aforementioned problems and the non-convexity of $\mathrm{SO}(p)$ render this problem difficult. 

\begin{figure}
    \centering
     \includegraphics[width=\linewidth, trim={0.4cm 0 1.5cm 0.5cm},clip]{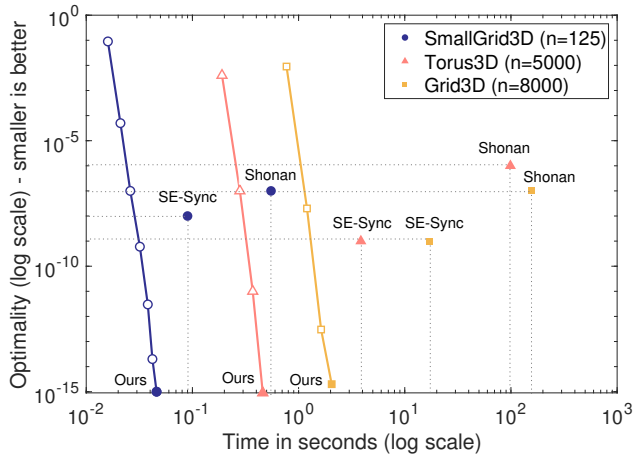}
    \caption{Comparison between our primal-dual method and the solutions produced by Shonan Averaging \cite{Dellaert2020} and SE-Sync \cite{Rosen2019} for rotation averaging problems adapted from pose graph optimization datasets \cite{Carlone2015}.}
    \vspace{-0.2cm}
    \label{fig:convergence_comparison}
\end{figure}

Under the hypothesis of the Langevin noise model \cite{Boumal2014,Chen2021} adopted in \cite{Eriksson2018,Dellaert2020,Arrigoni2020,Moreira2021} we formalize rotation averaging as the Maximum Likelihood Estimation (MLE) problem
\begin{equation}
    \begin{aligned}
        &\underset{R_1,\dots,R_n}{\textrm{minimize}} & & \sum_{i\sim j} \big\vert\big\vert \widetilde{R}_{ij} - R_i R_j^\top \big\vert\big\vert_F^2 \\
        &\textrm{subject to} & &  R_i \in \mathrm{SO(3)}, \;\; i = 1, \ldots, n.
    \end{aligned}
 \label{eq:intro_first}
\end{equation}
\paragraph{Contribution} Firstly, we present a primal-dual method to solve (\ref{eq:intro_first}) inspired in optimization algorithms with orthogonality constraints \cite{gao2019}. We show empirically that this algorithm converges to the global optimum when the dual variable is initialized with the graph degree matrix (Fig. \ref{fig:convergence_comparison}). Secondly, we put forward the first optimal closed-form solution for rotation averaging problems with a cycle graph topology. This solution allows for the retrieval of machine-precision global optima several orders of magnitude faster than the state-of-the-art and compounds the results in spectral graph theory set forth in \cite{Eriksson2018}. Our code is available at \url{https://github.com/gabmoreira/maks}.


\section{Related work}
The literature on rotation averaging spans a large array of optimization schemes, from iterative search methods \cite{Aloise2020,Dellaert2020,Eriksson2018,Eriksson2019,Nasiri2020,Rosen2019} to closed-form suboptimal solutions \cite{Arrigoni2020,Martinec2007,Moreira2021}. Recently, global optimality has taken the spotlight in rotation averaging papers. Nevertheless, finding the global minimum and thus solving (\ref{eq:intro_first}) remains a difficult task to accomplish efficiently. 

In the domain of iterative algorithms, Gauss-Newton (GN) and Levenberg-Marquardt (LM) methods such as those available in the pose graph optimization frameworks g2o \cite{Kummerle2011} and GTSAM \cite{Dellaert2012} were until recently, the most prominent techniques for solving rotation averaging. The non-convexity of this problem however, makes these techniques initialization dependent. A distributed Riemannian gradient descent algorithm in the manifold of 3D rotations has been proposed by Tron \etal \cite{Tron2009} but this method is arguably less efficient than GN and LM. 

In order to circumvent the retrieval of local optima, which are of no practical interest for the applications we are considering, Eriksson \etal \cite{Eriksson2018} derived the dual for problem (\ref{eq:intro_first}) and set forth a multitude of results pertaining to optimality verification relying on duality theory. Assuming strong duality holds, the solution of the semidefinite program (SDP) corresponding to the dual of problem (\ref{eq:intro_first}) yields the sought-after set of rotations. To solve this SDP, the authors presented a block coordinate descent method.

Alternative globally optimal strategies have been put forward. The SE-Sync pose graph optimization framework by Rosen \etal \cite{Rosen2019} relies on a Riemannian staircase \cite{boumal2015, boumal2016} to solve a SDP relaxation and guarantees globally optimal solutions under a few assumptions on the noise model. Shonan averaging, proposed by Dellaert \etal \cite{Dellaert2020} and now part of GTSAM, solves the problem locally on SO(3) and then increases the dimension of the manifold to start the optimization again. This is carried out iteratively, until a globally optimal solution is attained. This method combines the performance of GN and LM methods with a strategy for guaranteeing global optimality.

As opposed to the methods aforementioned, a number of works have proposed closed-form approximations of the optimal solution. Martinec \etal \cite{Martinec2007} treat the problem as a least-squares and then project the solution to the space of rotations. More recently, Arrigoni \etal \cite{Arrigoni2016, Arrigoni2020} and Moreira \etal \cite{Moreira2021} have proposed eigenspace-based solutions attending to the fact that for noise-free measurements solving rotation averaging is tantamount to solving an eigenvector equation. Whilst these closed-form approaches may yield satisfactory results for moderate noise levels, it is difficult to ascertain their domain of applicability.

\section{Problem statement}
Let $\mathcal{G}=(\mathcal{V},\mathcal{E})$ be a connected graph and $\{\widetilde{R}_{ij}\}_{i \sim j} \in \mathrm{SO}(3)^m$ a set of relative rotation measurements between nodes $i$ and $j$. Under the assumption of isotropic Langevin noise \cite{Boumal2014,Chen2021}, rotation averaging seeks the set of rotations $\{R_i^\ast\}_{i \in \mathcal{V}}$, which minimize the chordal distance \cite{Hartley2011} between each measurement $\widetilde{R}_{ij}$ and the respective pairwise estimate $R_i^\ast R_j^{\ast\top}$, over all edges of the graph. The MLE estimate \cite{Carlone2015, Moreira2021, Dellaert2020} is
\begin{equation}
    \{R_i^\ast\}_{i \in \mathcal{V}} = \argmax_{\{R_i\}_{i \in \mathcal{V}} \in \mathrm{SO(3)}^n} \sum_{i \sim j} \Tr (R_i^\top \widetilde{R}_{ij} R_j).
    \label{eq:rotavg2}
\end{equation}
We now introduce a block-matrix notation that we will use throughout the paper. Let $\mathrm{S}^p$ be the set of symmetric $p\times p$ matrices, $I_p \in \mathbb{R}^{p\times p}$ the identity and $0_p \in \mathbb{R}^{p\times p}$ the null matrix. We define the block-vector $R \in \mathrm{SO}(3)^n$ as
\begin{equation}
    R := \begin{bmatrix}
    R_1^\top &
    \dots &
    R_n^\top
    \end{bmatrix}^\top
\end{equation}
and the pairwise block-matrix $\widetilde{R}\in \mathrm{S}^{3n}$ as
\begin{equation}
    \widetilde{R}:=\begin{cases}
        I_3 &\quad \mathrm{if}\; i=j \\
        \widetilde{R}_{ij} \in \mathrm{SO}(3) &\quad \mathrm{if}\; i \sim j \\
        0_3 &\quad \mathrm{if}\; i \not\sim j.
    \end{cases}
    \label{eq:def_rtilde}
\end{equation}
The block-entry $i,j$ of $\widetilde{R}$ contains the  measured rotation of that edge or a null block if $i \not\sim j$. We set rotations from each node to itself as the identity and consider $\widetilde{R}_{ji} = \widetilde{R}_{ij}^\top$. Defining the cost function
\begin{equation}
f(R) := -\Tr \big(R^\top \widetilde{R} R\big),
\label{eq:def_f}
\end{equation}
we can write the optimization problem in (\ref{eq:rotavg2}) as
\begin{equation}
    \begin{aligned}
        &\underset{R}{\text{minimize}} & & f(R) \\
        &\text{subject to} & & R \in \mathrm{SO(3)}^n.
    \end{aligned}
\label{eq:rotation_averaging_problem}
\end{equation}

\section{Primal-dual method}
In this section, we present a novel primal-dual update method to solve problem (\ref{eq:rotation_averaging_problem}). We will show in Section \ref{sec:results} that in the applications considered, this algorithm succeeds in retrieving the global optimum. 

As derived in \cite{Eriksson2018}, the Lagrangian for rotation averaging under the orthogonality constraint $R\in \mathrm{O}(3)^n$ is
\begin{equation}
    \mathcal{L}(R,\Lambda) = -\Tr\big(R^\top\widetilde{R}R\big) - \Tr\big( \Lambda(I - R R^\top)\big),
    \label{eq:rotation_averaging_lagrangian}
\end{equation}
\noindent where the Lagrange multiplier is the block diagonal matrix $\Lambda = \mathrm{blockdiag}(\Lambda_1,\dots,\Lambda_n)$, with $\Lambda_i \in \mathrm{S}^3$. Differentiating (\ref{eq:rotation_averaging_lagrangian}), we have the stationarity condition
\begin{equation}
    \big(\Lambda - \widetilde{R}\big)R = 0
    \label{eq:rotation_averaging_stationarity}
\end{equation}
i.e., the optimal rotations are in the kernel of $\Lambda - \widetilde{R}$. Conversely, we can obtain $\Lambda$ from (\ref{eq:rotation_averaging_stationarity}) via
\begin{equation}
    \Lambda_i = R_i R_i^\top + \sum_{i \sim j} \widetilde{R}_{ij} R_j R_i^\top, \;\;\;\; i=1,\dots,n.
    \label{eq:lagrange_multiplier_formula}
\end{equation}
Our primal-dual method consists of combining (\ref{eq:rotation_averaging_stationarity}) and (\ref{eq:lagrange_multiplier_formula}) with projections to $\mathrm{SO}(3)^n$ and $\mathrm{S}^{3n}$ respectively, in order to create primal and dual feasible update rules.

\paragraph{Primal update} Given an estimate of the dual variable at the $k$-th iteration, which we denote by $\Lambda^k$, Eq. (\ref{eq:rotation_averaging_stationarity}) will in general not have a solution in $\mathrm{SO}(3)^n$. We resort thus to an approximation. In order to avoid the trivial solution $R=0$, we look for $X$ on the Stiefel manifold
\begin{equation}
    \mathrm{St}(3n,3) := \{X \in \mathbb{R}^{3n\times 3} : X^\top X = I_3 \}
\end{equation}
that minimizes $\Tr\big(X^\top(\Lambda^k - \widetilde{R})X\big)$. We then project the result to $\text{SO}(3)^n$ by solving $n$ Procrustes problems \cite{Schonemann1966}. Our primal updates consist thus of
\begin{align}
    X^{k+1} &= \argmin_{X \in \mathrm{St}(3n,3)} \mathrm{tr}\big(X^\top(\Lambda^k - \widetilde{R})X\big) \label{eq:problem_in_stiefel} \\
    R^{k+1} &= \argmin_{R \in \mathrm{SO}(3)^n} \big\vert\big\vert R - X^{k+1} \big\vert\big\vert_F^2 . \label{eq:procrustes}
\end{align}
The solution of (\ref{eq:problem_in_stiefel}) is given by the three eigenvectors of $\Lambda^k - \widetilde{R}$ associated with the three smallest eigenvalues. These eigenspaces can be computed efficiently by means of sparse symmetric eigensolvers. The optimization problem in (\ref{eq:procrustes}) can be solved via singular value decompositions of $3\times 3$ matrices. As demonstrated in \cite{Eriksson2018}, if a primal-dual pair $(R^\ast,\Lambda^\ast)$ verifies the stationarity condition (\ref{eq:rotation_averaging_stationarity}), then $\Lambda^\ast - \widetilde{R} \succeq 0$ is sufficient for strong duality to hold and for $(R^\ast,\Lambda^\ast)$ to be optimal. Thus, (\ref{eq:problem_in_stiefel}) allows for an optimality assessment at each iteration.

\paragraph{Dual update} Drawing from the work of Gao \etal on optimization problems with orthogonality constraints \cite{gao2019}, we form the dual update by symmetrizing  (\ref{eq:lagrange_multiplier_formula}). Let $\Psi(X)$ denote the projection to $\mathrm{S}^3$ \ie, $\Psi(X):=\frac{1}{2}(X+X^\top)$. Given a primal variable estimate $R^k$, we compute $\Lambda$ according to
\begin{align}
    \Lambda_i^{k} &= R_i^{k} R_i^{{k}^\top} + \Psi\bigg(\sum_{j \sim i} \widetilde{R}_{ij} R_{j}^{k} R_i^{{k}^\top}\bigg), i=1,\dots,n
\end{align}

\begin{figure}
    \centering
    \includegraphics[width=\linewidth, trim={0.2cm 0 1.4cm 0.7cm},clip]{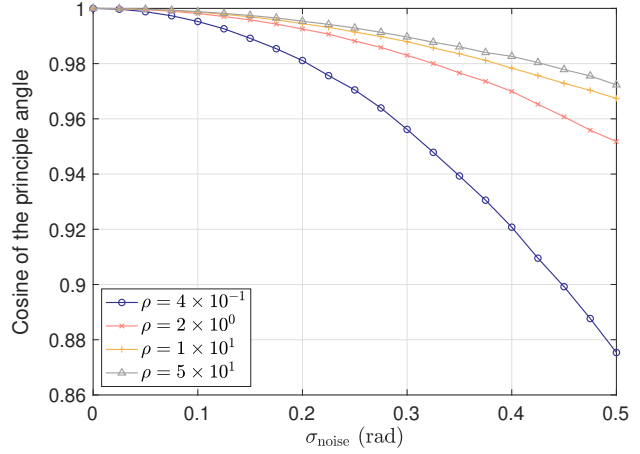}
    \caption{Cosine of the principal angle between the noisy and noise-free eigenspaces of $\Lambda_{\mathrm{nf}}-\widetilde{R}$ for a graph with 50 nodes and different algebraic connectivities $\rho$ (Fiedler value). Averaged over $3\times 10^4$ simulations.}
    \label{fig:principal_angle}
\end{figure}

\paragraph{Initialization} Instead of initializing the aforementioned primal-dual updates with an estimate of the primal variable i.e., a set of rotation estimates, we leverage the fact that for noise-free measurements $\Lambda^\ast$ depends only on the graph topology. In this case, the optimal rotations $R^\ast$ verify
\begin{equation}
    \big((D+I) \otimes I_3 - \widetilde{R}\big)R^\ast = 0,
\end{equation}
where $D\in\mathbb{R}^{n\times n}$ is the graph degree matrix. The optimal Lagrange multiplier for noise-free measurements is thus
\begin{equation}
   \Lambda_{\mathrm{nf}} := (D+I) \otimes I_3,
\end{equation}
as stated in \cite{Arrigoni2020,Moreira2021}. In our method, we set $\Lambda^0 = \Lambda_\mathrm{nf}$. To understand why this initialization allows the primal-dual iterations to attain optimality, we show empirically that for moderate noise levels, the subspace containing the ground-truth rotations and the subspace which solves (\ref{eq:problem_in_stiefel}) at $k=0$ are close together. Let $\widetilde{R}_\mathrm{nf} \in \mathrm{S}^{3n}$ denote the ground-truth pairwise block-matrix and $\widetilde{R}\in\mathrm{S}^{3n}$  (\ref{eq:def_rtilde}) the matrix obtained by perturbing the non-null blocks of $\widetilde{R}_\mathrm{nf}$ with Langevin noise (standard deviation $\sigma_\mathrm{noise}$). We represent in Fig. \ref{fig:principal_angle} the cosine of the principal angle between the kernel of $\Lambda_{\mathrm{nf}} - \widetilde{R}_\mathrm{nf}$, which we denote by $U_\mathrm{nf}$ and the subspace spanned by the three eigenvectors of $\Lambda_{\mathrm{nf}} - \widetilde{R}$ associated with the smallest eigenvalues, which we denote by $U$. The cosine is computed according to $\cos^2=\Tr\big(U^\top U_\mathrm{nf} U_\mathrm{nf}^\top U\big)/3$. As expected, these subspaces are closer for well connected graphs (large Fiedler value $\rho$). However, even for poorly connected ones the cosine is on average close to 1.

In Algorithm \ref{algo:rotavg} we show how our primal-dual updates were implemented. This method will henceforth be referred to as Rotation Averaging in a Split Second (RAveSS). The parameter $\sigma$ used in the sparse eigensolver corresponds to the eigenvalue target for the eigenvectors we are computing. Since our primal update is achieved by solving (\ref{eq:problem_in_stiefel}), we pick $\sigma < 0$ such that the three eigenvectors retrieved correspond to the three smallest eigenvalues. Note that prior to projecting the solution of the primal problem to $\text{SO}(3)^n$ we fix the gauge freedom by anchoring the first rotation.

\begin{algorithm}
\SetAlgoLined
\KwResult{$R, \Lambda$}
$\Lambda \gets (D+I_n)\otimes I_3$\;
\For {$t\gets 0$ to $\text{maxiter}$}{
    $X, \lambda_1, \lambda_2, \lambda_3 \gets \text{eigensolver}\big(\Lambda - \widetilde{R}, \sigma=-10^{-6}\big)$\;
    $X \gets X X_1^{-1}$\;
    \For {$i\gets 1$ to $n$}{
        $U \Sigma V^\top \gets \text{SVD}\big(X_i\big)$\;
        $R_i \gets U \;\text{diag}\big([1\; 1 \; \text{det}(UV^\top)]\big)\; V^\top$\;
    }
    \For {$i\gets 1$ to $n$}{
        $\Lambda_i \gets \frac{1}{2}\sum_{j\sim i}\widetilde{R}_{ij}R_jR_i^\top + \frac{1}{2}\big(\sum_{j\sim i}\widetilde{R}_{ij}R_jR_i^\top\big)^\top + R_i R_i^\top$\;
    }
    $\Lambda \gets \text{blockdiag}(\Lambda_1,\dots,\Lambda_n)$\;
    \If{$\min\{\vert\lambda_1\vert, \vert\lambda_2\vert, \vert\lambda_3\vert\} < \epsilon$}{
        \Return $R, \Lambda$\;
    }
}
\caption{RAveSS (Primal-Dual)}
\label{algo:rotavg}
\end{algorithm}

\section{Cycle graphs: optimal closed-form solution}
From a topological standpoint, cycle graphs are the simplest instance of rotation averaging. Nevertheless, they are usually solved via the same iterative methods that cater for the general synchronization problem \cite{Eriksson2018, Dellaert2020, Rosen2019}. In this section we show that rotation averaging problems with an underlying cycle graph topology have closed-form solutions for their stationary points, global optima included. We will first derive the closed-form solutions for one-parameter subgroups of SO(3). We then show that, in the general case, there is a basis wherein the block matrix $\widetilde{R}$ can be written such that its non-null blocks lie in a one-parameter subgroup of SO(3). This allows us to optimally solve problem (\ref{eq:rotation_averaging_problem}).

\paragraph{Cycle error} We start by defining the error $E \in \mathrm{SO}(3)$ incurred while traversing the cycle graph starting and ending on the same node. Without loss of generality let
\begin{equation}
    E := \prod_{k=0}^{n-1} \widetilde{R}_{\mathrm{mod}(k,n)+1,\;\mathrm{mod}(k+1,n)+1},
    \label{eq:def_cycle_error}
\end{equation}
with the matrix product in (\ref{eq:def_cycle_error}) being defined from left to right. In a cycle with 3 nodes \eg, $E = \widetilde{R}_{12}\widetilde{R}_{23}\widetilde{R}_{31}$. 

Further, let $\gamma \in [-\pi,\pi]$ be the angle of $E$, which we denote by $\gamma := \angle(E)$. We define the set of the $n$-th roots of $E$ as
\begin{equation}
    E^{\frac{1}{n}} := \big\{E_0,E_1,\dots,E_{n-1}\big\},
\end{equation}
with $E_k \in \mathrm{SO}(3)$, $E_k^n = E$ and $\angle(E_k) = \gamma/n - 2k\pi/n$, for  $k \in \{0,\dots,n-1\}$.

\subsection{One-parameter subgroups of SO(3)}
\label{sec:optimization_in_so2}
We consider for now one-parameter subgroups of SO(3) by assuming that the pairwise rotation measurements $\widetilde{R}_{12},\widetilde{R}_{23},\dots,\widetilde{R}_{n1}$ share a common axis.

\begin{lemma} For cycle graphs whose edge measurements lie in a one-parameter subgroup of SO(3), the points
\begin{equation}
    R_i = \Bigg(\prod_{s=1}^{i-1} \widetilde{R}_{s,s+1}\Bigg)^\top E_k^{i-1},\quad i\in\{2,\dots,n\}
    \label{eq:stationary_solution_cycle_so2}
\end{equation}
with $R_1 = I_3$, indexed by $k\in\{0,\dots,n-1\}$, are stationary points of problem (\ref{eq:rotation_averaging_problem}).

\begin{proof}
We rewrite the cost function $f$ (\ref{eq:def_f}) as
\begin{equation}
    f(R)=-3n - 2\sum_{i\sim j}\Tr\big(\widetilde{R}_{ij}R_jR_i^\top\big).
    \label{eq:trace}
\end{equation}
Under the hypothesis that the rotations $\widetilde{R}_{ij}$ share a common axis, we can restrict our search for $R_1,\dots,R_n$ to this subgroup. Thus, $\angle(\widetilde{R}_{ij}R_jR_i^\top)=\angle(\widetilde{R}_{ij}) - \angle(R_i R_j^\top)$. From $\Tr(R) = 1 + 2\cos(\angle(R))$, the trace in (\ref{eq:trace}) becomes
\begin{align}
    \Tr\big(\widetilde{R}_{ij}R_jR_i^\top\big) = 1+2\cos\big(\angle(\widetilde{R}_{ij}) - \angle(R_i R_j^\top)\big).
\end{align}
Define the angles $\theta_{ij}:=\angle(R_i R_j^\top)$, $\widetilde{\theta}_{ij}:=\angle(\widetilde{R}_{ij})$ and the set $\Theta=\{\theta_{12},\dots,\theta_{n1}\}$. The optimization problem
\begin{equation}
    \begin{aligned}
        &\underset{\Theta}{\textrm{maximize}} & & \sum_{i \sim j} \cos\big(\widetilde{\theta}_{ij} - \theta_{ij}\big) \\
        &\textrm{subject to} & &  \theta_{12} + \dots + \theta_{n1} = 2k\pi,
    \end{aligned}
    \label{eq:cosine_problem}
\end{equation}
for $k\in\{0,\dots,n-1\}$, is equivalent to (\ref{eq:rotation_averaging_problem}). Let the residuals be $\widetilde{\theta}_{ij} - \theta_{ij} \in [-\pi, \pi]$ and let $y \in \mathbb{R}$ be a dual variable. The Lagrangian for (\ref{eq:cosine_problem}) is
\begin{align}
    \mathcal{L}(\Theta, y) = \sum_{i \sim j} \cos(\widetilde{\theta}_{ij} - \theta_{ij}) + y \Bigg(\sum_{i\sim j}\theta_{ij} -2k\pi\Bigg)
    \label{eq:cos_lagrangian}
\end{align}
From (\ref{eq:cos_lagrangian}),  we have the sufficient stationarity conditions
\begin{align}
    &\exists \ w\in [-\pi,\pi],\; \forall_{i\sim j} \ \widetilde{\theta}_{ij}-\theta_{ij}  = w,  \label{eq:kktcos3}\\
    &\exists \ k \in \{0,\dots,n-1\}, \; \sum_{i\sim j}\theta_{ij} = 2k\pi.
    \label{eq:kktcos4}
\end{align}
Summing (\ref{eq:kktcos3}) over all the edges of the cycle graph we have
\begin{equation}
    \sum_{i\sim j}\widetilde{\theta}_{ij} - \sum_{i\sim j}\theta_{ij} = n w.
    \label{eq:kktcos5}
\end{equation}
Combining (\ref{eq:kktcos4}) and (\ref{eq:kktcos5}) with $\sum_{i\sim j}\widetilde{\theta}_{ij}=\gamma$ yields
\begin{align}
\theta_{ij} = \widetilde{\theta}_{ij} - \gamma/n + 2k\pi/n.
\label{eq:f_optimal_thetas}
\end{align}
From (\ref{eq:stationary_solution_cycle_so2}), we have $R_i R_j^\top = \widetilde{R}_{ij}E_k^\top$, for $i\sim j$. Thus, 
\begin{equation}
    \angle\big(R_i R_j^\top\big) = \angle\big(\widetilde{R}_{ij}\big) - \angle\big(E_k\big)
\end{equation}
which is simply (\ref{eq:f_optimal_thetas}) since $\angle(E_k) = \gamma/n - 2k\pi / n$.
\end{proof}
\label{prop:solution_in_so2}
\end{lemma}

\begin{theorem}
For cycle graphs whose edge measurements lie in a one-parameter subgroup of SO(3), the point
\begin{equation}
    R_i^\ast = \Bigg(\prod_{k=1}^{i-1} \widetilde{R}_{k,k+1}\Bigg)^\top E_0^{i-1},\quad i\in\{2,\dots,n\}
    \label{eq:optimal_solution_cycle_so2}
\end{equation}
with $R_1^\ast = I_3$ is a solution of problem (\ref{eq:rotation_averaging_problem}).
\begin{proof}
Invoking Theorem 4.2 of \cite{Eriksson2018} for cycle graphs, strong duality will hold and a solution will be globally optimal if $\forall i\sim j$ the residuals verify $\big\vert \widetilde{\theta}_{ij} - \theta^\ast_{ij} \big\vert \leq \frac{\pi}{n}$. From (\ref{eq:optimal_solution_cycle_so2}), the optimal rotations verify
\begin{equation}
    R^\ast_{j} =  E_0\widetilde{R}_{ij}^\top R^\ast_{i},
    \label{eq:recursion_optimal}
\end{equation}
with $\angle(E_0)=\gamma/n$. From (\ref{eq:recursion_optimal}) we can write
\begin{equation}
    \widetilde{\theta}_{ij} - \theta^\ast_{ij} = \angle\big(\widetilde{R}_{ij} R_{j}^\ast R_{i}^{\ast\top}\big) = \gamma/n,
\end{equation}
Since
\begin{equation}
    \vert \gamma /n \vert \leq \frac{\pi}{n}
\end{equation}
due to $\gamma \in [-\pi, \pi]$, the solution in (\ref{eq:optimal_solution_cycle_so2}) is optimal.
\end{proof}
\label{prop:global_min_so2}
\end{theorem}
In cycle graphs, rotation averaging problems in one-parameter subgroups of SO(3) will redistribute the cycle error equitably over all of the edges. If we incur an error of $E$, with $\angle(E)=\gamma$, the optimal relative rotation $R_i^\ast R_j^{\ast\top}$ will have an angular residual of $\gamma/n$ relative to the respective measurement $\widetilde{R}_{ij}$. By increasing this figure by a multiple of $2\pi / n$ we obtain suboptimal stationary points of (\ref{eq:rotation_averaging_problem}).


\subsection{Optimization in SO(3)}
We now show that any cycle graph problem in SO(3) has the same expression for its stationary points (\ref{eq:stationary_solution_cycle_so2}) and global optimum (\ref{eq:optimal_solution_cycle_so2}) as derived for one-parameter subgroups. We accomplish this by rewriting $\widetilde{R}$ in a new basis.
\paragraph{Change-of-basis} Define the matrix $U \in \mathrm{SO}(3n)$ as
\begin{equation}
     U := \mathrm{blockdiag}\big(U_1,\dots,U_n\big),
     \label{eq:defu}
\end{equation}
with $U_i \in \mathrm{SO}(3)$ for $i \in \{1,\dots,n\}$ computed according to
\begin{equation}
U_i :=
\begin{cases}
    I_3, \quad & i=1 \\
    \widetilde{R}_{i-1, i}^\top U_{i-1}, \quad & i\in\{2,\dots,n\}.
    \label{eq:defu2}
\end{cases}
\end{equation}

\begin{lemma} Denote by $\widetilde{R}'$, the matrix $\widetilde{R}$ written in the basis $U$ i.e., $\widetilde{R}' := U^\top \widetilde{R} U$. Then,
\begin{align}
    \widetilde{R}' = \begin{bmatrix}
    I_3 & I_3 & \dots & 0 & E^\top \\
    I_3 & I_3 & \dots & 0 & 0 \\
    \vdots & \vdots & \ddots & \vdots & \vdots\\
    0 & 0 & \dots & I_3 & I_3 \\
    E & 0 & \dots & I_3 & I_3 \\
    \end{bmatrix},
    \label{eq:Rtildeprime}
\end{align}
with $E=\widetilde{R}_{12}\widetilde{R}_{23}\dots \widetilde{R}_{n1}$ being the cycle error (\ref{eq:def_cycle_error}).
\begin{proof}
From (\ref{eq:defu2}), the blocks on the lower triangular part of $\widetilde{R}':=U^\top \widetilde{R}U$ are given by
\begin{equation}
    \widetilde{R}'_{ij} = 
    \begin{cases}
            U_n^\top \widetilde{R}_{n1} U_1, & i=n, j=1 \\
            U_i^\top \widetilde{R}_{ij} \widetilde{R}_{ij}^\top U_i, & j=i-1 \\
            U_i^\top U_j, & i=j.
    \end{cases}
    \label{eq:blocks_in_bases}
\end{equation}
It is immediate that for $i=j$ and $j=i-1$ we have $\widetilde{R}'_{ij}=I_3$. It suffices to show that $U_n^\top \widetilde{R}_{n1} U_1 = E$. Note that from (\ref{eq:defu2}) we have $U_1=I_3$ and $U_n = \widetilde{R}_{n-1,n}^\top \dots \widetilde{R}_{23}^\top \widetilde{R}_{12}^\top $. Thus,
\begin{equation}
    U_n^\top \widetilde{R}_{n1} U_1  = \widetilde{R}_{12}\widetilde{R}_{23}\dots \widetilde{R}_{n-1,n}\widetilde{R}_{n1},
    \label{eq:un_replaced}
\end{equation}
which equals $E$ by definition of the latter.
\end{proof}
\label{prop:cast_so3_so2}
\end{lemma}

\begin{figure}
\centering
\input{equivalent_graphs.tikz}
\vspace{-0.3cm}
\caption{The cycle graph problem on the left can be transformed into the problem on the right via a change-of-basis.}
\label{fig:equivalence}
\end{figure}
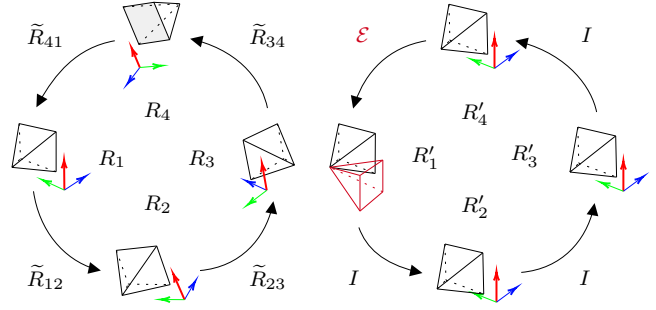

We can visualize this result in Fig. \ref{fig:equivalence}. In cycle graphs, MLE rotation averaging (\ref{eq:rotation_averaging_problem}) can be solved by concentrating the cycle error $E$ at a single edge. Further, by changing basis, the pairwise measurements $I_3$ and $E$ belong to a one-parameter subgroup of SO(3). We can thus leverage the results from Section \ref{sec:optimization_in_so2} to retrieve the global optimum and stationary points of problem (\ref{eq:rotation_averaging_problem}) in closed-form.

\begin{theorem} For cycle graphs with edge measurements in SO(3), the point
\begin{align}
    {R}_i^\ast &= \Bigg(\prod_{s=1}^{i-1} {\widetilde{R}}_{s,s+1} \Bigg)^\top E_0 ^{i-1},\quad i\in\{2,\dots,n\}
\end{align}
with $R_1^\ast= I_3$, is a solution of the problem (\ref{eq:rotation_averaging_problem}).
\label{prop:optimal_in_so3}

\begin{proof}
We write $f$ as
\begin{equation}
    f(R) = \Tr\big((U^\top R)^\top U^\top \widetilde{R} U (U^\top R)\big).
\end{equation}
Using the change-of-variables $R' = U^\top R$ and the change-of-basis $\widetilde{R}' = U^\top \widetilde{R} U$ we have the equivalent problem
\begin{equation}
    \begin{aligned}
        &\underset{R'}{\textrm{minimize}} & & \Tr\big({R'}^\top \widetilde{R}' R'\big) \\
        &\textrm{subject to} & &  R' \in \mathrm{SO}(3)^n,
    \end{aligned}
    \label{eq:newbasisproblem}
\end{equation}
whose edge measurements are $I_3$ and $E$ according to Lemma \ref{prop:cast_so3_so2}. These rotations belong to the one-parameter subgroup $t\mapsto \exp\big(t[\hat{n}]_\times\big)$, where $\hat{n}$ is the axis of $E$. Theorem \ref{prop:global_min_so2} is thus applicable and the solution of (\ref{eq:newbasisproblem}) is
\begin{equation}
    {R'}_i^\ast =  E_0^{i-1}, \quad i\in\{1,\dots,n\},
    \label{eq:solution_in_so2}
\end{equation}
since $\widetilde{R}'_{i,i+1}=I_3, \; \forall i\in\{1,\dots,n-1\}$. It suffices now to write (\ref{eq:solution_in_so2}) in the old basis vectors according to $R^\ast=U {R'}^\ast$. Since $U$ is block-diagonal, $R_i^\ast = U_i {R'}_i^\ast$. From the definition of $U_i$ (\ref{eq:defu2}) we have
\begin{equation}
    {R}_i^\ast = \Bigg(\prod_{s=1}^{i-1} {\widetilde{R}}_{s,s+1} \Bigg)^\top E_0 ^{i-1},\quad i\in\{2,\dots,n\}
\end{equation}
with ${R}_1^\ast = I_3$.
\end{proof} 
\end{theorem}
As a corollary of Theorem \ref{prop:optimal_in_so3}, we can take any stationary point of the problem in the new basis (see Lemma \ref{prop:solution_in_so2}) and revert to the old basis vectors in order to obtain the corresponding stationary point of problem (\ref{eq:rotation_averaging_problem}). Thus, the points
\vspace{-0.3cm}
\begin{equation}
    R_i = \Bigg(\prod_{s=1}^{i-1} {\widetilde{R}}_{s,s+1} \Bigg)^\top E_k ^{i-1},\quad i\in\{2,\dots,n\}
    \label{eq:evalf}
\end{equation}
with $R_1=I_3$, are stationary points of (\ref{eq:rotation_averaging_problem}) indexed by $k\in\{0,\dots,n-1\}$, where the cost function evaluates to
\begin{equation}
    f(R) = -3n-2n \Tr\big(E_k\big).
    \label{eq:stationary_law}
\end{equation}
Since $\Tr\big(E_k\big)=1+2\cos(\gamma/n-2k\pi/n)$ it follows that the greater the number of nodes, the greater the number of local minima near the global optimum. Hence the difficulty of solving rotation averaging optimally. 

We conclude this section by showing that in cycle graphs the spectrum of $\widetilde{R}$ relates to the values of $f$ at stationary points and can therefore be computed in closed-form.
\begin{theorem}
Let $\sigma(\widetilde{R})$ denote the spectrum of $\widetilde{R}$. Then,
\begin{align}
    \sigma\big(\widetilde{R}\big) = &\big\{1 + 2\cos\big(\gamma/n - 2k\pi/n\big) \big\}_{k=0,\dots, n-1}\nonumber \\
    &\cup \big\{1 + 2\cos\big(2k\pi/n\big)\big\}_{k=0,\dots, n-1}.
\end{align}
\label{prop:spectrum_in_so3}
The proof is provided in the appendix (Section \ref{sec:appendix}).
\end{theorem}


\begin{table*}[h]
\small
\begin{center}
\setlength{\tabcolsep}{5.5pt} 
\renewcommand{\arraystretch}{1.1}
\begin{tabular}{|l|cc|crc|ccr|ccr|}
\hline
& \multicolumn{2}{c|}{Graph} & \multicolumn{3}{c|}{RAveSS (ours)} & \multicolumn{3}{c|}{Shonan Averaging} &  \multicolumn{3}{c|}{SE-Sync} \\
Dataset & $n$  & $m$ & $\vert\lambda_0\vert$ & $f^\ast $ & $t (s)$ &  $\vert\lambda_0\vert$ & $\delta^\ast$ (approx.) & $t (s)$  & $\vert\lambda_0\vert$ & $\delta^\ast$ (approx.) & $t (s)$ \\
\hline\hline
SmallGrid  & $125$ & $297$ & $\mathbf{10^{-15}}$ & $\mathbf{-2118.202}$ & $\mathbf{0.02}$ & $10^{-07}$ & $-10^{-04}$ & 0.55 & $10^{-08}$ &   $-10^{-05}$  & 0.09 \\
Garage    & $1661$ & $6275$ & $\mathbf{10^{-15}}$ & $\mathbf{-42632.998}$  & $\mathbf{0.06}$  & $10^{-05}$ & $-10^{-01}$ & 24.5 & $10^{-14}$ & $-10^{-10}$ & 0.99\\
Sphere    & $2200$ & $8647$ & $\mathbf{10^{-15}}$ & $\mathbf{-56981.692}$ & $\mathbf{0.36}$ & $10^{-07}$ & $-10^{-03}$ & 30.1  & $10^{-09}$ & $-10^{-05}$ & 2.79 \\
Torus3D   & $5000$ & $9048$ & $\mathbf{10^{-15}}$ & $\mathbf{-69227.058}$ & $\mathbf{0.35}$ & $10^{-06}$ & $-10^{-02}$ & 98.8 & $10^{-09}$ & $-10^{-05}$ & 3.86 \\	
Cubicle   & $5750$ & $12486$ &  $\mathbf{10^{-15}}$ & $\mathbf{-92163.079}$ & $\mathbf{0.46}$ & $10^{-05}$ &  $-10^{-02}$ & 96.8 & $10^{-08}$  & $-10^{-04}$ & 2.49 \\
Grid3D    & $8000$ & $22236$ & $\mathbf{10^{-15}}$ &  $\mathbf{-157206.257}$ & $\mathbf{1.78}$ & $10^{-07}$ &  $-10^{-02}$ & 154.54 & $10^{-09}$  & $-10^{-04}$ & 11.69 \\
Rim       & $10195$ & $22251$ &  $\mathbf{10^{-15}}$ &  $\mathbf{-164037.930}$ & $\mathbf{5.92}$  & $10^{-05}$  & $-10^{-01}$ &  221.63 & $10^{-12}$ & $-10^{-07}$ & 8.73\\
\hline
\end{tabular}
\end{center}
\caption{Comparison between RAveSS (Algorithm \ref{algo:rotavg}), Shonan Averaging \cite{Dellaert2012} and SE-Sync \cite{Rosen2019}. Datasets from \cite{Carlone2015}.}
\label{tab:slam3d_benchmark}
\end{table*}

\section{Experimental results}
\label{sec:results}

In this section, we evaluate the performance of our primal-dual update method (RAveSS) and our closed-form solution in pose graph datasets and synthetic rotation averaging problems in cycle graphs, respectively. Our algorithms were implemented in C++ and all the tests were conducted on a laptop computer with a 6-core Intel Core i7-9750H@2.6GHz running macOS Big Sur.

\subsection{Pose graph datasets}
\label{sec:pgo_experiments}
Using seven datasets from the pose graph optimization literature available online \cite{Carlone2015}, we extracted the pairwise rotation measurements from each one in order to generate rotation averaging problems. Some of these datasets contain multiple measurements per edge, from which only one was kept. We compare the performance of RAveSS (primal-dual update method in Algorithm \ref{algo:rotavg}) against Shonan Averaging (SA) \cite{Dellaert2020} and SE-Sync \cite{Rosen2019}. The authors' implementations are available online and we tested them with their default parameters. Since SE-Sync is designed for solving pose graph optimization problems, we set the input translations to zero. The stopping criterion for our method was defined as $\vert\lambda_0\vert < 10^{-15}$, which corresponds to tolerance of the Krylov-based eigensolver used.

The results can be observed in Table \ref{tab:slam3d_benchmark}. In order to juxtapose the three methods in terms of the positive semidefiniteness of $\Lambda - \widetilde{R}$ \ie, in order to verify optimality, we proceeded as follows. For each estimate computed, we obtained the Lagrange multiplier using the KKT condition in (\ref{eq:lagrange_multiplier_formula}) and symmetrized it via the projection $\Psi$. The columns $\lambda_0$ in Table \ref{tab:slam3d_benchmark} correspond to the minimum eigenvalue of $\Psi(\Lambda) - \widetilde{R}$, which is zero if a given solution is optimal and strong duality holds. In addition, we represent the cost function evaluated at the solution produced by RAveSS, denoted by $f^\ast$ and the difference between this minimum and the minima computed by SA and SE-Sync, denoted by $\delta^\ast = f^\ast - f$. The CPU time is shown in seconds for all three algorithms.

The three methods benchmarked reach the global optimum in all seven datasets. While there may be disparities regarding precision, the differences in terms of the minimum attained and the set of rotations produced are negligeable in the applications considered. We focus thus on the CPU time of each algorithm. Our primal-dual method attains machine precision of $\lambda_0$, and therefore the global optimum, faster than the two other methods take to stop iterating. If we were to relax the upper bound on our stopping criterion, the CPU times could be brought down even further, without compromising the solution as far as geometric reconstructions are concerned. Plots showcasing the convergence of RAveSS in terms of the positive-semidefinitess of $\Lambda-\widetilde{R}$ for six of the datasets are available in the supplemental material.

\subsection{Cycle graphs}
Borrowing the evaluation approach adopted in \cite{Dellaert2020, Eriksson2018}, we tested our closed-form cycle graph solution, entitled C-RAveSS, in synthetic cycle graph datasets. These consisted of random rotation averaging problems with underlying cycle graph structures of different sizes, wherein the ground-truth absolute orientations $R_i$ correspond to rotations around the z-axis, forming a circular trajectory. The synthetic pairwise measurements were simulated by perturbing the relative ground-truth orientations between adjacent nodes by an error matrix obtained from angle-axis representations. The axes were sampled uniformly over the unit sphere. The angles were drawn from a normal distribution with zero mean and standard deviation $\sigma$. 

We benchmarked C-RAveSS against two baselines, the block coordinate descent method (BCD) \cite{Eriksson2018} used to solve the dual of the dual of problem (\ref{eq:rotation_averaging_problem}) and the SA algorithm which we also tested in our pose graph experiments. We implemented the former in MATLAB and used the author's implementation of the latter. Both methods were initialized randomly. Results averaged over 20 simulations can be observed in Table \ref{tab:cycle_benchmark}. For our solution, we list the smallest eigenvalue of $\Lambda - \widetilde{R}$, denoted by $\vert\bar{\lambda}_0\vert$, which certifies that, as we have shown, our solution is optimal to machine precision in all the simulations we ran. In the two rightmost columns, we show the average difference $\bar{\delta}^\ast$ between our closed-form global minimum and the cost function evaluated at the set of rotations produced by SA and BCD.

Using its default settings, SA retrieved the global optimum in all the tests conducted. Nevertheless, not only does precision wane as the order of the cycle increases but also the average CPU time surges substantially as the number of variables increases. In order to test the BCD method, we first computed the global optimum in each simulation with C-RAveSS. We then used it to set the stopping criterion for the BCD as $\delta^\ast \leq 10^{-3}$. As evidenced by the average CPU time, convergence flatlined for the largest cycles. Attaining the global minimum to three decimal places using this algorithm took, on average, as much as 96 seconds for $n=100$ and longer than that would be required for $n = 200$. While this may be a shortcoming of our implementation, the orders of magnitude of the CPU time appear to be in accordance with those reported in \cite{Dellaert2020}.

\begin{table}
\small
\begin{center}
\setlength{\tabcolsep}{4.2pt} 
\renewcommand{\arraystretch}{1.1}
\begin{tabular}{|cc|cr|cr|cr|}
\hline
\multicolumn{2}{|c|}{Problem} & \multicolumn{2}{c|}{C-RAveSS} & \multicolumn{2}{c|}{Shonan} &  \multicolumn{2}{c|}{BCD} \\
$n$  & $\sigma$ (rad) & $\vert \lambda_0 \vert$ & $\bar{t}$ (s) & $\bar{\delta}^\ast$ & $\bar{t}$ (s) & $\bar{\delta}^\ast$ & $\bar{t}$  (s)\\
\hline\hline
$20$ & $0.2$ & $\mathbf{10^{-15}}$  &  \textbf{0.00007} & $10^{-4}$ & 0.11 & $10^{-3}$  & 0.18 \\
 & $0.5$ & $\mathbf{10^{-15}}$  &  \textbf{0.00007} &  $10^{-4}$ & 0.12 & $10^{-3}$ &  0.23 \\

$50$ & $0.2$ & $\mathbf{10^{-15}}$  &  \textbf{0.00008} & $10^{-3}$ & 0.26 & $10^{-3}$ &  4.48 \\
 & $0.5$ & $\mathbf{10^{-15}}$  &  \textbf{0.00008} & $10^{-2}$ & 0.32 & $10^{-3}$ &   6.80 \\

$100$ & $0.2$ & $\mathbf{10^{-15}}$  & \textbf{0.00009} &  $10^{-2}$ & 0.42 & $10^{-3}$ & 51.75 \\
 & $0.5$ & $\mathbf{10^{-15}}$  & \textbf{0.00009}  & $10^{-2}$ & 0.50 & $10^{-3}$  &  96.19 \\

$200$ & $0.2$ & $\mathbf{10^{-15}}$  & \textbf{0.00010} & $10^{-2}$ &  0.74 & $10^{-3}$  & n.a.  \\
 & $0.5$ & $\mathbf{10^{-15}}$  & \textbf{0.00010} & $10^{-1}$  &  1.10 & $10^{-3}$  & n.a. \\
\hline
\end{tabular}
\end{center}
\caption{Comparison between our closed-form solution, Shonan Averaging \cite{Dellaert2020} and the block coordinate descent method (BCD) \cite{Eriksson2018} for random cycle graph problems.}
\label{tab:cycle_benchmark}
\end{table}

\begin{figure}
    \centering
    \includegraphics[width=\linewidth, trim={0.2cm 0.2cm 0.2cm 0.0cm},clip]{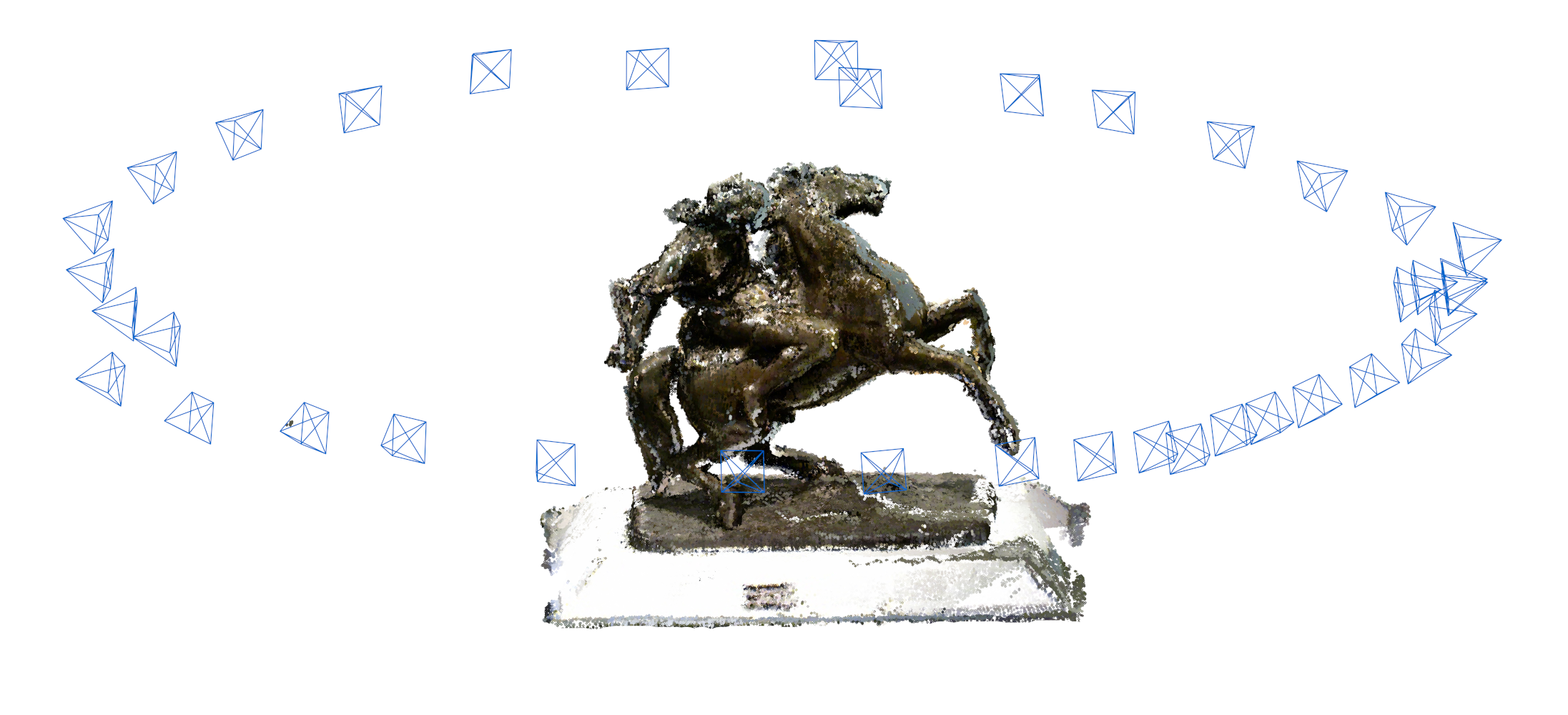}
    \caption{Reconstruction using our closed-form solution for cycle graphs with 38 optimal rotations computed in 70 $\mu s$.}
    \label{fig:horse}
\end{figure}

These experiments validate our solution as both the fastest and most precise method to solve cycle graphs. As per Table \ref{tab:cycle_benchmark}, our optimal closed-form represents a performance gain over the state-of-the-art that may be as big as 10000-fold. Fig. \ref{fig:horse} illustrates an application of this solution in geometric reconstruction using a cycle graph rotation averaging problem extracted from a larger 3D reconstruction dataset \cite{Choi2016}.

\section{Conclusion}

In this paper we presented two contributions to the problem of averaging multiple rotations. Considering the MLE formulation under the hypothesis of Langevin noise, we set forth a primal-dual update method and a closed-form solution for cycle graphs. As demonstrated by our empirical evaluation, the former produces optimal solutions to machine precision in a fraction of the time of existing solvers. Further, it verifies optimality by default at each iteration. The latter is, to our knowledge, the first optimal closed-form to be derived for this class of problems. The performance gain from optimally solving a problem that used to be tackled via iterative search methods, in closed-form, is substantial, as evidenced by the experiments conducted.

\section{Appendix}

\subsection{Proof of Theorem \ref{prop:spectrum_in_so3}}
\label{sec:appendix}
We will prove the result for the spectrum of $\widetilde{R}'$, which is equal to that of $\widetilde{R}$ since the matrices are similar. We start by showing that the block-vectors $V^k \in \mathbb{R}^{3n\times 3}$, with
\begin{equation}
    V^k := \begin{bmatrix}
    E_k^{0} \\ E_k^{1} \\ \vdots \\ E_k^{{n-1}}
    \end{bmatrix},
    \label{eq:vk}
\end{equation}
indexed by $k\in\{0,\dots,n-1\}$ span invariant subspaces of $\widetilde{R}'$ i.e., $\exists\; H \in \mathbb{R}^{3\times 3} : \widetilde{R}' V^k = V^k H$. For this step, it suffices to compute $\widetilde{R}' V ^k$. From (\ref{eq:Rtildeprime}) and (\ref{eq:vk})
\begin{align}
    \widetilde{R}'
    \begin{bmatrix}
    E_k^0 \\
    \vdots \\
    E_k^i \\
    \vdots \\
    E_k^{n-1}
    \end{bmatrix} &= 
    \begin{bmatrix}
    E_k^0 + E_k^1 + E^\top E_k^{n-1} \\
    \vdots \\
    E_k^{i-1} + E_k^i + E_k^{i+1} \\
    \vdots \\
    E E_k^0 + E_k^{n-2} + E_k^{n-1} \\
    \end{bmatrix} \nonumber \\
    &= \begin{bmatrix}
    E_k^0 \\
    \vdots \\
    E_k^i  \\
    \vdots \\
    E_k^{n-1} \\
    \end{bmatrix}\big(I_3 + E_k + E_k^\top\big).
    \label{eq:teo5_1}
\end{align}
It compact notation, (\ref{eq:teo5_1}) reads as 
\begin{equation}
    \widetilde{R}' V^k = V^k (I_3 + E_k + E_k^\top).
    \label{eq:compactinvarsubspace}
\end{equation}
Since $I_3 + E_k + E_k^\top \in \mathrm{S}^3$, let its EVD be
\begin{equation}
    I_3 + E_k + E_k^\top \stackrel{\mathrm{EVD}}{=} J \Sigma J^\top, 
    \label{eq:spectruminvarsub}
\end{equation}
with $J\in \mathrm{O}(3)$ and $\Sigma\in\mathbb{R}^{3\times 3}$ diagonal. From (\ref{eq:compactinvarsubspace}) we have
\begin{equation}
    \widetilde{R}' (V^k J) = (V^k J) \Sigma.
\end{equation}
The diagonal of $\Sigma$ contains thus three eigenvalues of $\widetilde{R}'$. From (\ref{eq:spectruminvarsub}), these eigenvalues are those of $I_3+E_k+E_k^\top$ i.e., $\{3, 1+2\cos(\angle(E_k)), 1+2\cos(\angle(E_k))\}$.  By definition, $\angle(E_k)=\gamma/n - 2k\pi/n$, thus
\begin{equation}
    \big\{1+2\cos(\gamma/n - 2k\pi/n)\big\}_{k=0,\dots,n-1} \subset \sigma(\widetilde{R}),
    \label{eq:theo5_spec1}
\end{equation}
with each eigenvalue having multiplicity 2.

In order to identify the remaining $n$ eigenvalues of $\widetilde{R}$ let $\hat{n}$ denote the axis of the cycle error $E$ i.e., $E\hat{n}=\hat{n}$ and $E^\top \hat{n} = \hat{n}$. Define the vectors $z^k\in\mathbb{R}^{3n}$
\begin{equation}
    z^k := \begin{bmatrix}
    1 \\
    \cos\big(1\frac{2k\pi}{n}\big) \\
    \cos\big(2\frac{2k\pi}{n}\big) \\
    \vdots \\
    \cos\big((n-1)\frac{2k\pi}{n}\big)
    \end{bmatrix} \otimes \hat{n},
\end{equation}
indexed by $k\in\{0,\dots,n-1\}$. We have
\begin{align}
    &\widetilde{R}' z^k = 
      \nonumber \\
    &\begin{bmatrix}
    \big(1 + \cos\big(\frac{2k\pi}{n}\big) + \cos\big((n-1)\frac{2k\pi}{n})\big)\hat{n}  \\
    \vdots \\
    \big(\cos\big((i-1)\frac{2k\pi}{n}\big)+ \cos\big(i\frac{2k\pi}{n}\big) + \cos\big((i+1)\frac{2k\pi}{n}\big) \big)\hat{n} \\
    \vdots \\
    \big(1 + \cos\big((n-2)\frac{2k\pi}{n}\big) + \cos\big((n-1)\frac{2k\pi}{n}\big) \big)\hat{n}
    \end{bmatrix} \nonumber \\
    &= \bigg(1+2\cos\bigg(\frac{2k\pi}{n}\bigg)\bigg)
    \begin{bmatrix}
    1\\
    \vdots \\
    \cos\big(i\frac{2k\pi}{n}\big) \\
    \vdots \\
    \cos\big((n-1)\frac{2k\pi}{n}\big) 
    \end{bmatrix} \otimes \hat{n}
    \label{eq:theo5compact_zk}
\end{align}
In compact notation, (\ref{eq:theo5compact_zk}) reads as
\begin{equation}
    \widetilde{R}' z^k = \big(1+2\cos(2k\pi/n)\big) z^k,
\end{equation}
for $k\in\{0,\dots,n-1\}$. It follows that 
\begin{equation}
    \big\{1+2\cos(2k\pi/n)\big\}_{k=0,\dots,n-1} \subset \sigma(\widetilde{R}).
    \label{eq:theo5_spec2}
\end{equation}
Finally, from (\ref{eq:theo5_spec1}) and (\ref{eq:theo5_spec2}) we have
\begin{align}
    \sigma(\widetilde{R}) = &\big\{1 + 2\cos\big(\gamma/n - 2k\pi/n\big) \big\}_{k=0,\dots, n-1}\nonumber \\
    &\cup \big\{1 + 2\cos\big(2k\pi/n\big)\big\}_{k=0,\dots, n-1}.
\end{align}

\paragraph{Acknowledgments} The authors would like to thank the reviewers for their comments and suggestions. This work was funded by Fundação para a Ciência e Tecnologia, grant [UIDB/50009/2020]. João Paulo Costeira and Manuel Marques were also supported by the European Union's Horizon 2020 project (GA 825619, AI4EU).

\newpage
{\small
\bibliographystyle{ieee_fullname}
\bibliography{egbib}
}


\end{document}

%% file: equivalent_graphs.tikz
\tikzset{every picture/.style={line width=0.3pt}} 

\begin{tikzpicture}[x=0.75pt,y=0.75pt,yscale=-1,xscale=1]

\draw    (121.66,93.92) -- (125.3,71.83) ;
\draw    (121.66,93.92) -- (143.92,75.96) ;
\draw    (125.3,71.83) -- (143.92,75.96) ;
\draw    (121.66,93.92) -- (143.92,96.12) ;
\draw    (143.92,75.96) -- (143.92,96.12) ;
\draw  [dash pattern={on 0.84pt off 2.51pt}]  (125.3,71.83) -- (125.95,89.55) ;
\draw  [dash pattern={on 0.84pt off 2.51pt}]  (121.66,93.92) -- (125.95,89.55) ;
\draw  [dash pattern={on 0.84pt off 2.51pt}]  (125.95,89.55) -- (143.92,96.12) ;
\draw  [fill={rgb, 255:red, 240; green, 240; blue, 240 }  ,fill opacity=1 ] (193.21,10.61) -- (201.13,29.48) -- (185.67,31.84) -- (177.75,12.97) -- cycle ;
\draw  [dash pattern={on 0.84pt off 2.51pt}]  (177.75,12.97) -- (206.31,11.34) ;
\draw    (193.21,10.61) -- (206.31,11.34) ;
\draw    (201.13,29.48) -- (206.31,11.34) ;
\draw  [dash pattern={on 0.84pt off 2.51pt}]  (185.67,31.84) -- (206.31,11.34) ;
\draw    (177.24,158.56) -- (174.3,135.81) ;
\draw    (177.24,158.56) -- (194.65,133.61) ;
\draw    (174.3,135.81) -- (194.65,133.61) ;
\draw    (177.24,158.56) -- (201.42,153.11) ;
\draw    (194.65,133.61) -- (201.42,153.11) ;
\draw  [dash pattern={on 0.84pt off 2.51pt}]  (174.3,135.81) -- (180.32,152.86) ;
\draw  [dash pattern={on 0.84pt off 2.51pt}]  (177.24,158.56) -- (180.32,152.86) ;
\draw  [dash pattern={on 0.84pt off 2.51pt}]  (180.32,152.86) -- (201.42,153.11) ;
\draw    (264.1,92.76) -- (255.53,71.92) ;
\draw    (264.1,92.76) -- (238.2,80.18) ;
\draw    (255.53,71.92) -- (238.2,80.18) ;
\draw    (264.1,92.76) -- (242.14,100.16) ;
\draw    (238.2,80.18) -- (242.14,100.16) ;
\draw  [dash pattern={on 0.84pt off 2.51pt}]  (255.53,71.92) -- (258.9,89.44) ;
\draw  [dash pattern={on 0.84pt off 2.51pt}]  (264.1,92.76) -- (258.9,89.44) ;
\draw  [dash pattern={on 0.84pt off 2.51pt}]  (258.9,89.44) -- (242.14,100.16) ;
\draw    (135.44,60.92) .. controls (143.32,46.85) and (154.2,33.4) .. (173.5,30.02) ;
\draw [shift={(133.97,63.61)}, rotate = 298.07] [fill={rgb, 255:red, 0; green, 0; blue, 0 }  ][line width=0.08]  [draw opacity=0] (6.25,-3) -- (0,0) -- (6.25,3) -- cycle    ;
\draw    (163.44,145.21) .. controls (147.69,140.45) and (135.32,125.74) .. (132.5,106) ;
\draw [shift={(166.5,146)}, rotate = 192.14] [fill={rgb, 255:red, 0; green, 0; blue, 0 }  ][line width=0.08]  [draw opacity=0] (6.25,-3) -- (0,0) -- (6.25,3) -- cycle    ;
\draw    (252.86,114.05) .. controls (249.02,129.87) and (237.18,144.12) .. (216.5,146) ;
\draw [shift={(253.5,111)}, rotate = 100.01] [fill={rgb, 255:red, 0; green, 0; blue, 0 }  ][line width=0.08]  [draw opacity=0] (6.25,-3) -- (0,0) -- (6.25,3) -- cycle    ;
\draw    (214.55,29.77) .. controls (238.33,36.3) and (248.13,51.89) .. (250.5,64) ;
\draw [shift={(211.5,29)}, rotate = 12.99] [fill={rgb, 255:red, 0; green, 0; blue, 0 }  ][line width=0.08]  [draw opacity=0] (6.25,-3) -- (0,0) -- (6.25,3) -- cycle    ;
\draw [color={rgb, 255:red, 255; green, 0; blue, 0 }  ,draw opacity=1 ][line width=0.75]    (147.95,92) -- (148.08,105.39) ;
\draw [shift={(147.93,90)}, rotate = 89.44] [color={rgb, 255:red, 255; green, 0; blue, 0 }  ,draw opacity=1 ][line width=0.75]    (4.37,-1.32) .. controls (2.78,-0.56) and (1.32,-0.12) .. (0,0) .. controls (1.32,0.12) and (2.78,0.56) .. (4.37,1.32)   ;
\draw [color={rgb, 255:red, 0; green, 15; blue, 255 }  ,draw opacity=1 ]   (148.08,105.39) -- (156.78,99.71) ;
\draw [shift={(158.45,98.62)}, rotate = 506.88] [color={rgb, 255:red, 0; green, 15; blue, 255 }  ,draw opacity=1 ][line width=0.75]    (4.37,-1.32) .. controls (2.78,-0.56) and (1.32,-0.12) .. (0,0) .. controls (1.32,0.12) and (2.78,0.56) .. (4.37,1.32)   ;
\draw [color={rgb, 255:red, 0; green, 255; blue, 15 }  ,draw opacity=1 ]   (139.13,101.24) -- (148.08,105.39) ;
\draw [shift={(137.32,100.4)}, rotate = 24.83] [color={rgb, 255:red, 0; green, 255; blue, 15 }  ,draw opacity=1 ][line width=0.75]    (4.37,-1.32) .. controls (2.78,-0.56) and (1.32,-0.12) .. (0,0) .. controls (1.32,0.12) and (2.78,0.56) .. (4.37,1.32)   ;
\draw [color={rgb, 255:red, 0; green, 15; blue, 255 }  ,draw opacity=1 ]   (250.14,105.52) -- (241.91,101.71) ;
\draw [shift={(240.1,100.87)}, rotate = 384.85] [color={rgb, 255:red, 0; green, 15; blue, 255 }  ,draw opacity=1 ][line width=0.75]    (4.37,-1.32) .. controls (2.78,-0.56) and (1.32,-0.12) .. (0,0) .. controls (1.32,0.12) and (2.78,0.56) .. (4.37,1.32)   ;
\draw [color={rgb, 255:red, 255; green, 0; blue, 0 }  ,draw opacity=1 ][line width=0.75]    (248.04,93.67) -- (250.14,105.52) ;
\draw [shift={(247.69,91.7)}, rotate = 79.96] [color={rgb, 255:red, 255; green, 0; blue, 0 }  ,draw opacity=1 ][line width=0.75]    (4.37,-1.32) .. controls (2.78,-0.56) and (1.32,-0.12) .. (0,0) .. controls (1.32,0.12) and (2.78,0.56) .. (4.37,1.32)   ;
\draw [color={rgb, 255:red, 0; green, 255; blue, 15 }  ,draw opacity=1 ]   (242.28,111.49) -- (250.14,105.52) ;
\draw [shift={(240.69,112.7)}, rotate = 322.78] [color={rgb, 255:red, 0; green, 255; blue, 15 }  ,draw opacity=1 ][line width=0.75]    (4.37,-1.32) .. controls (2.78,-0.56) and (1.32,-0.12) .. (0,0) .. controls (1.32,0.12) and (2.78,0.56) .. (4.37,1.32)   ;
\draw [color={rgb, 255:red, 0; green, 15; blue, 255 }  ,draw opacity=1 ]   (186.14,43.52) -- (181.18,50.02) ;
\draw [shift={(179.97,51.61)}, rotate = 307.31] [color={rgb, 255:red, 0; green, 15; blue, 255 }  ,draw opacity=1 ][line width=0.75]    (4.37,-1.32) .. controls (2.78,-0.56) and (1.32,-0.12) .. (0,0) .. controls (1.32,0.12) and (2.78,0.56) .. (4.37,1.32)   ;
\draw [color={rgb, 255:red, 255; green, 0; blue, 0 }  ,draw opacity=1 ][line width=0.75]    (182.74,35.46) -- (186.14,43.52) ;
\draw [shift={(181.97,33.61)}, rotate = 67.19] [color={rgb, 255:red, 255; green, 0; blue, 0 }  ,draw opacity=1 ][line width=0.75]    (4.37,-1.32) .. controls (2.78,-0.56) and (1.32,-0.12) .. (0,0) .. controls (1.32,0.12) and (2.78,0.56) .. (4.37,1.32)   ;
\draw [color={rgb, 255:red, 0; green, 255; blue, 15 }  ,draw opacity=1 ]   (186.14,43.52) -- (194.98,42.78) ;
\draw [shift={(196.97,42.61)}, rotate = 535.21] [color={rgb, 255:red, 0; green, 255; blue, 15 }  ,draw opacity=1 ][line width=0.75]    (4.37,-1.32) .. controls (2.78,-0.56) and (1.32,-0.12) .. (0,0) .. controls (1.32,0.12) and (2.78,0.56) .. (4.37,1.32)   ;
\draw [color={rgb, 255:red, 0; green, 15; blue, 255 }  ,draw opacity=1 ]   (208.69,160.7) -- (212.96,153.34) ;
\draw [shift={(213.97,151.61)}, rotate = 480.17] [color={rgb, 255:red, 0; green, 15; blue, 255 }  ,draw opacity=1 ][line width=0.75]    (4.37,-1.32) .. controls (2.78,-0.56) and (1.32,-0.12) .. (0,0) .. controls (1.32,0.12) and (2.78,0.56) .. (4.37,1.32)   ;
\draw [color={rgb, 255:red, 255; green, 0; blue, 0 }  ,draw opacity=1 ][line width=0.75]    (203.77,149.45) -- (208.69,160.7) ;
\draw [shift={(202.97,147.61)}, rotate = 66.39] [color={rgb, 255:red, 255; green, 0; blue, 0 }  ,draw opacity=1 ][line width=0.75]    (4.37,-1.32) .. controls (2.78,-0.56) and (1.32,-0.12) .. (0,0) .. controls (1.32,0.12) and (2.78,0.56) .. (4.37,1.32)   ;
\draw [color={rgb, 255:red, 0; green, 255; blue, 15 }  ,draw opacity=1 ]   (198.97,160.63) -- (208.69,160.7) ;
\draw [shift={(196.97,160.61)}, rotate = 0.41] [color={rgb, 255:red, 0; green, 255; blue, 15 }  ,draw opacity=1 ][line width=0.75]    (4.37,-1.32) .. controls (2.78,-0.56) and (1.32,-0.12) .. (0,0) .. controls (1.32,0.12) and (2.78,0.56) .. (4.37,1.32)   ;
\draw    (282,93.89) -- (286.48,71.95) ;
\draw    (282,93.89) -- (304.93,76.8) ;
\draw    (286.48,71.95) -- (304.93,76.8) ;
\draw    (282,93.89) -- (304.8,97.03) ;
\draw    (304.93,76.8) -- (304.8,97.03) ;
\draw  [dash pattern={on 0.84pt off 2.51pt}]  (286.48,71.95) -- (286.45,89.68) ;
\draw  [dash pattern={on 0.84pt off 2.51pt}]  (282,93.89) -- (286.45,89.68) ;
\draw  [dash pattern={on 0.84pt off 2.51pt}]  (286.45,89.68) -- (304.8,97.03) ;
\draw [color={rgb, 255:red, 255; green, 0; blue, 0 }  ,draw opacity=1 ][line width=0.75]    (366.01,148.61) -- (366.01,161.97) ;
\draw [shift={(366.01,146.61)}, rotate = 90] [color={rgb, 255:red, 255; green, 0; blue, 0 }  ,draw opacity=1 ][line width=0.75]    (4.37,-1.32) .. controls (2.78,-0.56) and (1.32,-0.12) .. (0,0) .. controls (1.32,0.12) and (2.78,0.56) .. (4.37,1.32)   ;
\draw [color={rgb, 255:red, 0; green, 15; blue, 255 }  ,draw opacity=1 ]   (366.01,161.97) -- (374.36,155.8) ;
\draw [shift={(375.97,154.61)}, rotate = 503.57] [color={rgb, 255:red, 0; green, 15; blue, 255 }  ,draw opacity=1 ][line width=0.75]    (4.37,-1.32) .. controls (2.78,-0.56) and (1.32,-0.12) .. (0,0) .. controls (1.32,0.12) and (2.78,0.56) .. (4.37,1.32)   ;
\draw [color={rgb, 255:red, 0; green, 255; blue, 15 }  ,draw opacity=1 ]   (356.83,158.35) -- (366.01,161.97) ;
\draw [shift={(354.97,157.61)}, rotate = 21.52] [color={rgb, 255:red, 0; green, 255; blue, 15 }  ,draw opacity=1 ][line width=0.75]    (4.37,-1.32) .. controls (2.78,-0.56) and (1.32,-0.12) .. (0,0) .. controls (1.32,0.12) and (2.78,0.56) .. (4.37,1.32)   ;
\draw    (336.07,155.28) -- (340.04,132.61) ;
\draw    (336.07,155.28) -- (359.13,137.06) ;
\draw    (340.04,132.61) -- (359.13,137.06) ;
\draw    (336.07,155.28) -- (359.58,157.84) ;
\draw    (359.13,137.06) -- (359.58,157.84) ;
\draw  [dash pattern={on 0.84pt off 2.51pt}]  (340.04,132.61) -- (340.52,150.83) ;
\draw  [dash pattern={on 0.84pt off 2.51pt}]  (336.07,155.28) -- (340.52,150.83) ;
\draw  [dash pattern={on 0.84pt off 2.51pt}]  (340.52,150.83) -- (359.58,157.84) ;
\draw    (403.07,96.28) -- (407.04,73.61) ;
\draw    (403.07,96.28) -- (426.13,78.06) ;
\draw    (407.04,73.61) -- (426.13,78.06) ;
\draw    (403.07,96.28) -- (426.58,98.84) ;
\draw    (426.13,78.06) -- (426.58,98.84) ;
\draw  [dash pattern={on 0.84pt off 2.51pt}]  (407.04,73.61) -- (407.52,91.83) ;
\draw  [dash pattern={on 0.84pt off 2.51pt}]  (403.07,96.28) -- (407.52,91.83) ;
\draw  [dash pattern={on 0.84pt off 2.51pt}]  (407.52,91.83) -- (426.58,98.84) ;
\draw [color={rgb, 255:red, 255; green, 0; blue, 0 }  ,draw opacity=1 ][line width=0.75]    (430.01,92.61) -- (430.01,105.97) ;
\draw [shift={(430.01,90.61)}, rotate = 90] [color={rgb, 255:red, 255; green, 0; blue, 0 }  ,draw opacity=1 ][line width=0.75]    (4.37,-1.32) .. controls (2.78,-0.56) and (1.32,-0.12) .. (0,0) .. controls (1.32,0.12) and (2.78,0.56) .. (4.37,1.32)   ;
\draw [color={rgb, 255:red, 0; green, 15; blue, 255 }  ,draw opacity=1 ]   (430.01,105.97) -- (438.36,99.8) ;
\draw [shift={(439.97,98.61)}, rotate = 503.57] [color={rgb, 255:red, 0; green, 15; blue, 255 }  ,draw opacity=1 ][line width=0.75]    (4.37,-1.32) .. controls (2.78,-0.56) and (1.32,-0.12) .. (0,0) .. controls (1.32,0.12) and (2.78,0.56) .. (4.37,1.32)   ;
\draw [color={rgb, 255:red, 0; green, 255; blue, 15 }  ,draw opacity=1 ]   (420.83,102.35) -- (430.01,105.97) ;
\draw [shift={(418.97,101.61)}, rotate = 21.52] [color={rgb, 255:red, 0; green, 255; blue, 15 }  ,draw opacity=1 ][line width=0.75]    (4.37,-1.32) .. controls (2.78,-0.56) and (1.32,-0.12) .. (0,0) .. controls (1.32,0.12) and (2.78,0.56) .. (4.37,1.32)   ;
\draw    (294.66,61.24) .. controls (300.08,45.66) and (311.2,33.4) .. (330.5,30.02) ;
\draw [shift={(293.69,64.26)}, rotate = 286.39] [fill={rgb, 255:red, 0; green, 0; blue, 0 }  ][line width=0.08]  [draw opacity=0] (6.25,-3) -- (0,0) -- (6.25,3) -- cycle    ;
\draw    (378.56,30.77) .. controls (402.49,37.38) and (414.13,53.89) .. (416.5,66) ;
\draw [shift={(375.5,30)}, rotate = 12.99] [fill={rgb, 255:red, 0; green, 0; blue, 0 }  ][line width=0.08]  [draw opacity=0] (6.25,-3) -- (0,0) -- (6.25,3) -- cycle    ;
\draw    (414.86,114.98) .. controls (411.02,130.81) and (399.18,145.06) .. (378.5,146.94) ;
\draw [shift={(415.5,111.94)}, rotate = 100.01] [fill={rgb, 255:red, 0; green, 0; blue, 0 }  ][line width=0.08]  [draw opacity=0] (6.25,-3) -- (0,0) -- (6.25,3) -- cycle    ;
\draw    (338.07,34.28) -- (342.04,11.61) ;
\draw    (338.07,34.28) -- (361.13,16.06) ;
\draw    (342.04,11.61) -- (361.13,16.06) ;
\draw    (338.07,34.28) -- (361.58,36.84) ;
\draw    (361.13,16.06) -- (361.58,36.84) ;
\draw  [dash pattern={on 0.84pt off 2.51pt}]  (342.04,11.61) -- (342.52,29.83) ;
\draw  [dash pattern={on 0.84pt off 2.51pt}]  (338.07,34.28) -- (342.52,29.83) ;
\draw  [dash pattern={on 0.84pt off 2.51pt}]  (342.52,29.83) -- (361.58,36.84) ;
\draw [color={rgb, 255:red, 255; green, 0; blue, 0 }  ,draw opacity=1 ][line width=0.75]    (365.01,30.61) -- (365.01,43.97) ;
\draw [shift={(365.01,28.61)}, rotate = 90] [color={rgb, 255:red, 255; green, 0; blue, 0 }  ,draw opacity=1 ][line width=0.75]    (4.37,-1.32) .. controls (2.78,-0.56) and (1.32,-0.12) .. (0,0) .. controls (1.32,0.12) and (2.78,0.56) .. (4.37,1.32)   ;
\draw [color={rgb, 255:red, 0; green, 15; blue, 255 }  ,draw opacity=1 ]   (365.01,43.97) -- (373.36,37.8) ;
\draw [shift={(374.97,36.61)}, rotate = 503.57] [color={rgb, 255:red, 0; green, 15; blue, 255 }  ,draw opacity=1 ][line width=0.75]    (4.37,-1.32) .. controls (2.78,-0.56) and (1.32,-0.12) .. (0,0) .. controls (1.32,0.12) and (2.78,0.56) .. (4.37,1.32)   ;
\draw [color={rgb, 255:red, 0; green, 255; blue, 15 }  ,draw opacity=1 ]   (355.83,40.35) -- (365.01,43.97) ;
\draw [shift={(353.97,39.61)}, rotate = 21.52] [color={rgb, 255:red, 0; green, 255; blue, 15 }  ,draw opacity=1 ][line width=0.75]    (4.37,-1.32) .. controls (2.78,-0.56) and (1.32,-0.12) .. (0,0) .. controls (1.32,0.12) and (2.78,0.56) .. (4.37,1.32)   ;
\draw [color={rgb, 255:red, 208; green, 2; blue, 27 }  ,draw opacity=1 ]   (282,93.89) -- (308.69,90.26) ;
\draw [color={rgb, 255:red, 208; green, 2; blue, 27 }  ,draw opacity=1 ]   (282,93.89) -- (297.17,98) ;
\draw [color={rgb, 255:red, 208; green, 2; blue, 27 }  ,draw opacity=1 ]   (308.69,90.26) -- (308.69,106.26) ;
\draw [color={rgb, 255:red, 208; green, 2; blue, 27 }  ,draw opacity=1 ]   (282,93.89) -- (297.17,116) ;
\draw [color={rgb, 255:red, 208; green, 2; blue, 27 }  ,draw opacity=1 ]   (308.69,106.26) -- (297.17,116) ;
\draw [color={rgb, 255:red, 208; green, 2; blue, 27 }  ,draw opacity=1 ]   (297.17,98) -- (297.17,116) ;
\draw [color={rgb, 255:red, 208; green, 2; blue, 27 }  ,draw opacity=1 ]   (308.69,90.26) -- (297.17,98) ;
\draw [color={rgb, 255:red, 208; green, 2; blue, 27 }  ,draw opacity=1 ] [dash pattern={on 0.84pt off 2.51pt}]  (308.69,106.26) -- (282,93.89) ;
\draw    (325.63,146.56) .. controls (308.63,142.35) and (300.38,134.63) .. (296.6,124.24) ;
\draw [shift={(328.69,147.26)}, rotate = 191.91] [fill={rgb, 255:red, 0; green, 0; blue, 0 }  ][line width=0.08]  [draw opacity=0] (6.25,-3) -- (0,0) -- (6.25,3) -- cycle    ;

\draw (164,84) node [anchor=north west][inner sep=0.75pt]  [font=\footnotesize]  {$R_{1}$};
\draw (187,107) node [anchor=north west][inner sep=0.75pt]  [font=\footnotesize]  {$R_{2}$};
\draw (209.3,84) node [anchor=north west][inner sep=0.75pt]  [font=\footnotesize]  {$R_{3}$};
\draw (187,60) node [anchor=north west][inner sep=0.75pt]  [font=\footnotesize]  {$R_{4}$};
\draw (128,142) node [anchor=north west][inner sep=0.75pt]  [font=\footnotesize]  {$\widetilde{R}_{12}$};
\draw (128,20) node [anchor=north west][inner sep=0.75pt]  [font=\footnotesize]  {$\widetilde{R}_{41}$};
\draw (240,20) node [anchor=north west][inner sep=0.75pt]  [font=\footnotesize]  {$\widetilde{R}_{34}$};
\draw (290,145) node [anchor=north west][inner sep=0.75pt]  [font=\footnotesize]  {$I$};
\draw (347,107) node [anchor=north west][inner sep=0.75pt]  [font=\footnotesize]  {$R'_{2}$};
\draw (372.01,80.97) node [anchor=north west][inner sep=0.75pt]  [font=\footnotesize]  {$R'_{3}$};
\draw (346.95,60) node [anchor=north west][inner sep=0.75pt]  [font=\footnotesize]  {$R'_{4}$};
\draw (406.55,145) node [anchor=north west][inner sep=0.75pt]  [font=\footnotesize]  {$I$};
\draw (407.55,22) node [anchor=north west][inner sep=0.75pt]  [font=\footnotesize]  {$I$};
\draw (294,22) node [anchor=north west][inner sep=0.75pt]  [font=\footnotesize,color={rgb, 255:red, 208; green, 2; blue, 27 }  ,opacity=1 ]  {$\mathcal{E}$};
\draw (322,82) node [anchor=north west][inner sep=0.75pt]  [font=\footnotesize]  {$R'_{1}$};
\draw (240,142) node [anchor=north west][inner sep=0.75pt]  [font=\footnotesize]  {$\widetilde{R}_{23}$};

\end{tikzpicture}